\documentclass[runningheads]{llncs}

\usepackage[T1]{fontenc}
\def\doi#1{\href{https://doi.org/\detokenize{#1}}{\url{https://doi.org/\detokenize{#1}}}}
\usepackage{graphicx}
\usepackage{url}            
\usepackage{booktabs}       
\usepackage{amsfonts}       
\usepackage{nicefrac}       
\usepackage{microtype}      
\usepackage{xcolor}         
\usepackage{listings}
\usepackage{algorithm} 
\usepackage{algpseudocode} 

 \usepackage[sort,comma,numbers]{natbib}

\begin{document}
\title{Time Majority Voting, a PC-based EEG Classifier for Non-expert Users}
\titlerunning{Time Majority Voting}
%
\author{Guangyao Dou\inst{1}\orcidID{0000-0001-8011-9658} \and
Zheng Zhou\inst{1}\orcidID{0000-0001-9313-2106} \and
Xiaodong Qu\inst{1,2}\orcidID{0000-0001-7610-6475}}

\authorrunning{G. Dou, Z. Zhou and X. Qu}
%
\institute{Brandeis University, Waltham MA 02453, USA\\ \email{\{guangyaodou,zhengzhou\}@brandeis.edu}
\and Swarthmore College, Swarthmore PA 19081, USA\\
\email{xqu1@swarthmore.edu}}
%
\maketitle              
\begin{abstract}
Using Machine Learning and Deep Learning to predict cognitive tasks from electroencephalography (EEG) signals is a rapidly advancing field in Brain-Computer Interfaces (BCI). In contrast to the fields of computer vision and natural language processing, the data amount of these trials is still rather tiny. Developing a PC-based machine learning technique to increase the participation of non-expert end-users could help solve this data collection issue. We created a novel algorithm for machine learning called Time Majority Voting (TMV). In our experiment, TMV performed better than cutting-edge algorithms. It can operate efficiently on personal computers for classification tasks involving the BCI. 
These interpretable data also assisted end-users and researchers in comprehending EEG tests better.
\end{abstract}
\section{Introduction}

Researchers from Computer Science, Neuroscience, and Medical fields have applied EEG-based Brain-Computer Interaction (BCI) techniques in many different ways \cite{lotte2018review,craik2019deep,qu2020identifying,appriou2020modern,lotte2010regularizing, lotte2014tutorial,kastrati2021eegeyenet}, such as diagnosis of abnormal states, evaluating the effect of the treatments, seizure detection, motor imagery tasks \cite{devlaminck2010circular,lotte2015signal,lotte2015towards,bashivan2014spectrotemporal,bashivan2015learning,bashivan2016mental}, and developing BCI-based games \cite{coyle2013guest}. Previous studies have demonstrated the great potential of machine learning, deep learning, and transfer learning algorithms \cite{lotte2007review,zhang2020survey,zhao2019bira,roy2019deep,miller2019library,kaya2018large,zhao2020sea,lotte2018bci,bird2018study,li2021hierarchical,gu2020multi,zhao2022adaptive,zeng2021robust,an2021mars,basaklar2021wearable, bhatwearable,derby2020computational,chen2021data} in such clinical and non-clinical data analysis.

However, the data size of such experiments is still relatively small compared to the areas of computer vision or natural language processing. Thus, some deep learning or big data approaches still struggling with the limitation of small dataset size. Also,
EEG signals have noise issues, partly because of the contact of sensors and skin for several current non-invasive consumer-grade devices. The outlier issue is also a concern for the EEG data because of the difficulties subjects have in concentrating on the experimental tasks during the entire session. Current machine learning and deep learning algorithms are more for clinical experiments and less for the possible experiment for non-expert user to conduct at home. 

Our research questions are: Can we develop a PC-based machine learning algorithm for non-expert end-users to do EEG classification at home? Can we achieve reasonably high accuracy while keeping the run time in an acceptable range? Can we make the machine learning classification results explainable to the end-users? To answer these questions, we proposed a new machine learning classification algorithm, Time Majority Voting (TMV). We found TMV outperformed other state-of-the-art classifiers. Also, its run time on a PC is still acceptable compared to the deep learning algorithms. The classification results are adequately interpretable to the end-users.

The paper is organized as follows: section two discusses several most frequently used classification algorithms for BCI research, then present our new algorithm. Section three presents our experiment conducted to test the new algorithm. Section four elaborates our result followed by sedition five, which discusses the limitation and future work. Lastly, section six concludes the study and summarized our answers to the research questions.

\section{Algorithms}
All of the code was run on a 2018 Macbook Pro with a 2.2GHz 6-core Intel Core i7 processor and with 16 GB of memory. The Python version is 3.8. The scikit-learn \cite{scikit-learn} version is 0.24.1. The PyTorch \cite{paszke2017automatic} version is 1.10. The code discussed in this paper is available online (\url{https://github.com/GuangyaoDou/Time_Majority_Voting}).
\subsection{Existing Algorithms}
u
We reviewed and implemented several machine learning algorithms commonly used in the field \cite{breiman1996bagging,breiman2001random,breiman2017classification,chevalier2020statistical}. For examples, Linear Classifiers, Nearest Neighbors, Decision Trees, and Ensemble Methods. 

\textbf{Linear Classifiers}: The Shrinkage Linear Discriminant Analysis (Shrinkage LDA) performed adequately on EEG datasets with simple tasks. The Support Vector Machine (SVM), effective in high dimensional spaces, performed reasonably well based on the previous research. These algorithms are simple to implement and are computationally efficient.

\textbf{Nearest Neighbor}: Such a classifier implements the K-Nearest Neighbor(KNN). KNN performs voting to determine an unseen dataset to one of the k nearest neighbors. The KNN performed pretty well compared to most other classifiers on the EEG dataset.

\textbf{Decision Tree}: The Decision Trees classifier is easy to understand, implement, and interpret. The decision tree creates a model that predicts the outcome of a data point based on decision rules. The computational cost is low and can handle both numerical and categorical data. However, it might overfit as trees are too complex.

\textbf{Ensemble Methods}: We used Random Forest and boosting. These are Ensemble Machine Learning algorithms that combine the predictions of several weaker learners and form more robust and more accurate predictions. These have been widely used in EEG-based experiments, and research \cite{lotte2018review,qu2020multi,qu2020using}. 

\textbf{Deep Learning}: We implemented CNN with ReLu, and RNN, especially LSTM, mainly using toolsets from the PyTorch platform. \cite{roy2019deep, craik2019deep, qu2020identifying}. These DLs performed very well on EEG datasets. However, we excluded these algorithms in this paper due to their high runtime. 

\subsection{Our New Algorithms}
This paper proposed a new voting approach based on the top two individual classifiers. Ensemble methods, especially boosting, bagging, and voting, have demonstrated excellent performance in previous research. \cite{lotte2018review, roy2019deep, craik2019deep, qu2018eeg, qu2020identifying} In EEG-based BCI classification research, the following voting methods have been investigated in several experiments \cite{lotte2018review, qu2020multi}: majority voting, weighted voting, and time continuity voting. Here we considered the advantages of both majority voting and time continuity voting and developed our new Time Majority Voting (TMV) algorithm.

 \begin{figure}[!b]
\centering
  \includegraphics[width=0.9999\textwidth]{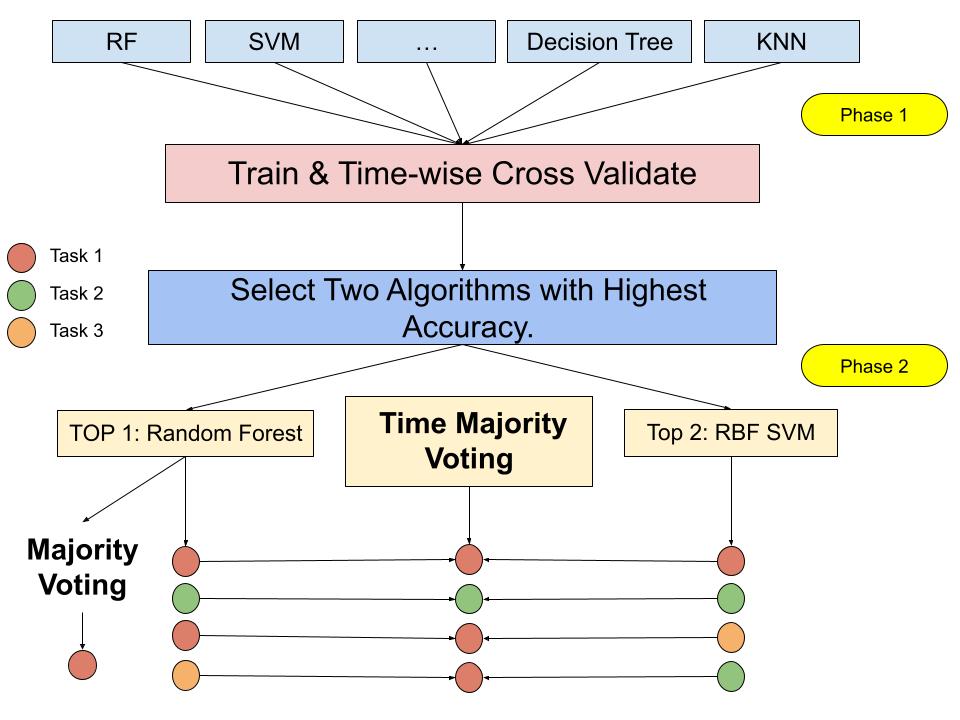}
  \caption{Our New Algorithm: Time Majority Voting}
  \label{fig:OurNewAlgorithm}
\end{figure}

\smallskip
\textbf{Time Majority Voting (TMV)}. ~\\
Figure \ref{fig:OurNewAlgorithm} demonstrated the concept of the new algorithm. More details are in the Experiment section and the Result section. There are two phases in the new algorithm. First, we investigated the state-of-the-art machine learning algorithms \cite{lotte2018review, roy2019deep, craik2019deep, qu2020identifying} and found the top two performers on average. In our experiments during phase 1, We tested Random Forest (RF), RBF and linear SVM, kNN, Decision Tree, and several boosting algorithms. We found Random Forest performed the best, and RBF SVM performed the second on average. Next, we entered phase two. For each subject, we picked the majority task predicted by the best performing classifier, the Random Forest classifier, for each time interval of each task from each session.

\smallskip
The next step is voting. We used the Random Forest and the RBF SVM to conduct the voting process. The best algorithm is the Random Forest, and the second algorithm is RBF SVM. The voting details are shown in the four examples at the bottom of Figure \ref{fig:OurNewAlgorithm}. If both the Random Forest and the RBF SVM algorithms agree with the results, as shown in the first two rows, the results reflect both algorithms' results. For example, as the second row shows, if both the classifiers predicted the task 2, then the Time Majority Voting will yield task 2 as the result. If the two algorithms do not agree on the prediction results, the results will be labeled as the majority of tasks already determined by the Random Forest classifier, as shown in the last two rows. No matter what the two algorithms predict out of the five tasks, even if they predicted it as two different tasks other than the majority task, as the last row shows, Task 2 and Task 3, the TMV result is still set to be Task one. This concept is based on the temporal dependence time-series features in the previous research \cite{lotte2007review, qu2018eeg}.

\section{Experiments}

Several EEG experiments focus on the high-level cognitive tasks that college students frequently conduct, as mentioned in Table \ref{tasks_experiment}. In this experiment, we used the dataset from the Think-Count-Recall (TCR) paper \cite{qu2020multi}. Scalp-EEG signals were recorded from seventeen subjects. Each one was tested in six sessions. Each session is five minutes long, with five tasks, each task is one minute. Tasks were selected by the subjects together with the researchers based on frequent tasks in study environments for students in their everyday lives. Each subject completed six sessions over several weeks. The five tasks are Think(T), Count(C), Recall(R), Breathe(B), and Draw(D). 

\begin{table}[!t]
\centering

\begin{tabular} { | c | c | c | c | c | c |}
\hline
(E) T & 1 & 2 & 3 & 4 & 5\\
\hline
\cite{qu2018personalized} & Math & Close-eye Relax & Read & Open-eye Relax & None \\
\hline
\cite{qu2018eeg} & Python Passive & Math Passive & Python Active & Math Active & None \\
\hline
\cite{qu2020using} & Read & Write Copy & Write Answer & Type Copy & Type Answer \\
\hline
\cite{qu2020multi} & Think & Count & Recall & Breathe & Draw\\
\hline
\end{tabular}
\vspace{.1in}
\caption{\label{tasks_experiment} Tasks (T) in Experiments (E)}
\end{table}

\subsection{Data Preprocess}

\textbf{Data cleaning:} As mentioned in \cite{qu2020using,qu2020multi}, for each task during each session, the first 30\% of the data, which is the first 18 seconds of each task and during the transition phase, will be removed. Thus, each one-minute task only had 42 seconds left. This has been proved reasonable during the data cleaning phase. Some electrodes may have temporarily lost contact with the subjects' scalp during the EEG recording. The result was that multiple sequential spectral snapshots from one or more electrodes had the same value. In this paper, we decide to remove such anomaly when detected for a consecutive 1.4 seconds. Such a cleaning action caused a different level of loss of the data for each subject. 

\textbf{Subjects:} We had a total of seventeen subjects. After the data cleaning actions, subjects who lost more than 65\% of the total data will be excluded from the subsequent analysis. In the end, there were twelve subjects left to continue the analysis. Moreover, for the six sessions of each subject, if a session lost more than 65\% of the data, then that session will also be excluded for further analysis. 

\textbf{Time-wise Cross-Validation:} We adopted time-wise cross-validation. We divided each task into seven subsets, meaning each subset had six seconds, evenly and continuously. Then, we created a total of seven folds. Each fold contained six seconds of data for each task for each session. We checked any folds that lost more than 65\% of the original data in each fold and discarded these folds for future analysis. Next, we used one fold for testing and the remaining non-discarded folds for training, and we cross-validated them.  

\section{Results}

\begin{table}[!b]
\centering
\begin{tabular} { | c | c | c |}
\hline
  Algorithms & Average Accuracy & Average code run-time (s) \\
\hline
    Random Forest Phase 1 & \textbf{0.55} & 42.0 \\
\hline
    RBF SVM & \textbf{0.53} & 30.5 \\ 
\hline
    Nearest Neighbors & 0.48 & \textbf{1.9} \\ 
\hline
    Decision Tree & 0.44 & 0.9 \\ 
\hline
    Linear SVM & 0.42 & 23.2 \\ 
\hline
    Shrinkage LDA & 0.42 & 0.2 \\ 
\hline
    Adaboost Classifier & 0.39 & 47.8 \\ 
\hline
    RUSBoost & 0.39 & 28.2 \\ 
\hline
    GradientBoost & 0.31 & 24.0 \\ 
\hline
\end{tabular}
\vspace{.1in}
\caption{\label{accuracy_runtime} State-of-the-art Algorithms with Accuracy and Run-time}
\end{table}

\subsection{Existing Algorithms}
We reported the average accuracy for all subjects and the runtime of each classifier for each subject we trained and tested during phase 1 in table \ref{accuracy_runtime}. As we can see, Random Forest had the highest accuracy of 0.55 in our experiment, and the SVM with RBF kernel performed adequately on the TCR dataset with an average accuracy of 0.53. Though Nearest Neighbors did not perform as well as the Random Forest and the RBF SVM, it was one of the fastest algorithms on personal computers. Other ensemble methods such as Adaboost, RusBoost, and GradientBoosting performed relatively lower than these top three algorithms.  

 \begin{figure}[!t]
\centering
  \includegraphics[width=0.9999\textwidth]{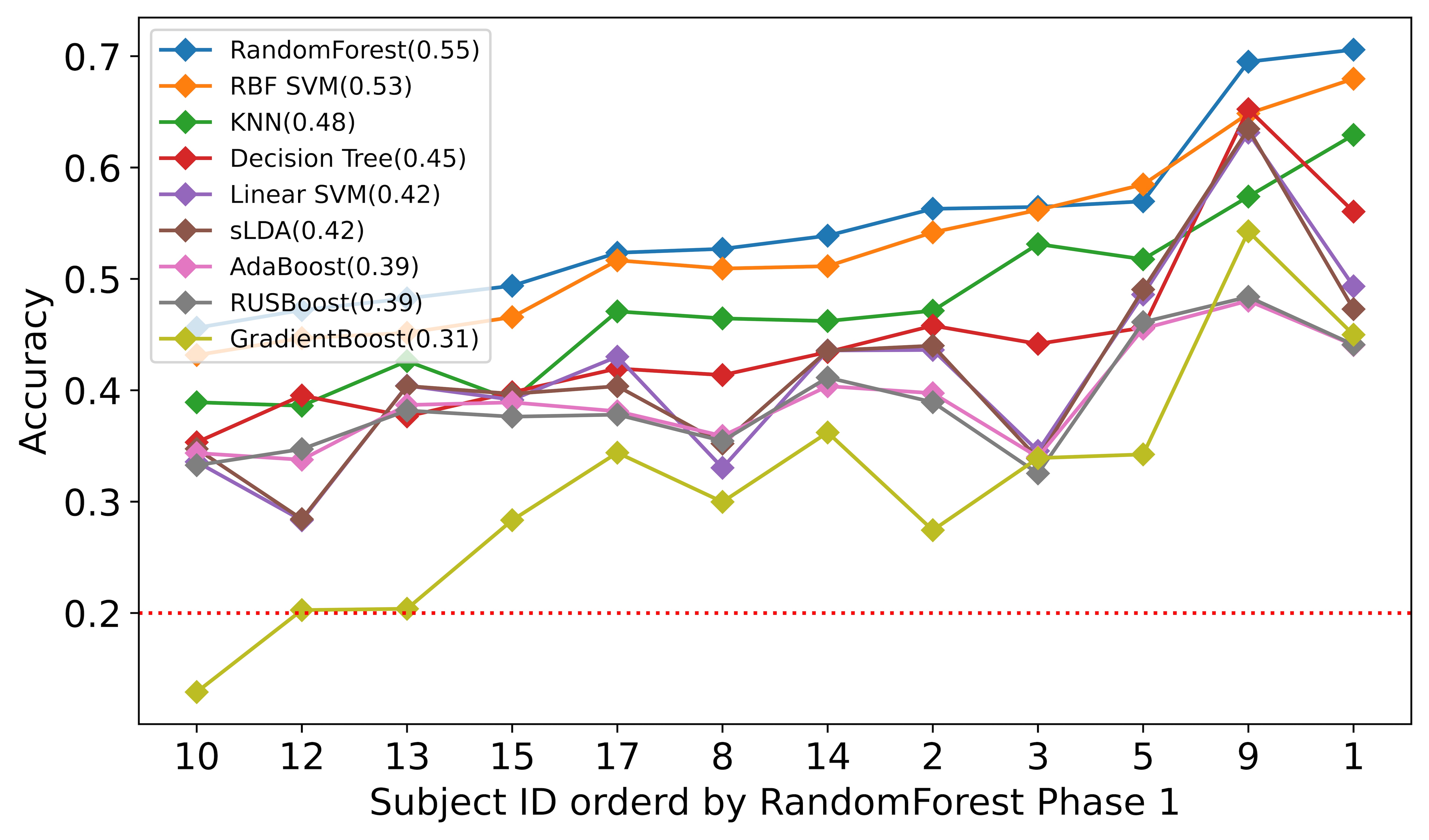}
  \caption{Accuracy for Different Algorithms}
  \label{fig:CompareAlgorithms}
\end{figure}

The individual difference may impact the accuracy of each subject. But we can still recognize a general pattern from Fig \ref{fig:CompareAlgorithms}. We ordered all twelve subjects by prediction accuracy using Random Forest. Most of the algorithms demonstrated consistent patterns for the different algorithms. Random Forest and RBF SVM are above most of the other algorithms. We kept the threshold of maintaining the subject to 35\% of the remaining data, as we believed that when a subject has little data left, the high accuracy from that subject contributes little to our research.  

\begin{figure}[!t]
\centering
  \includegraphics[width=0.999\textwidth]{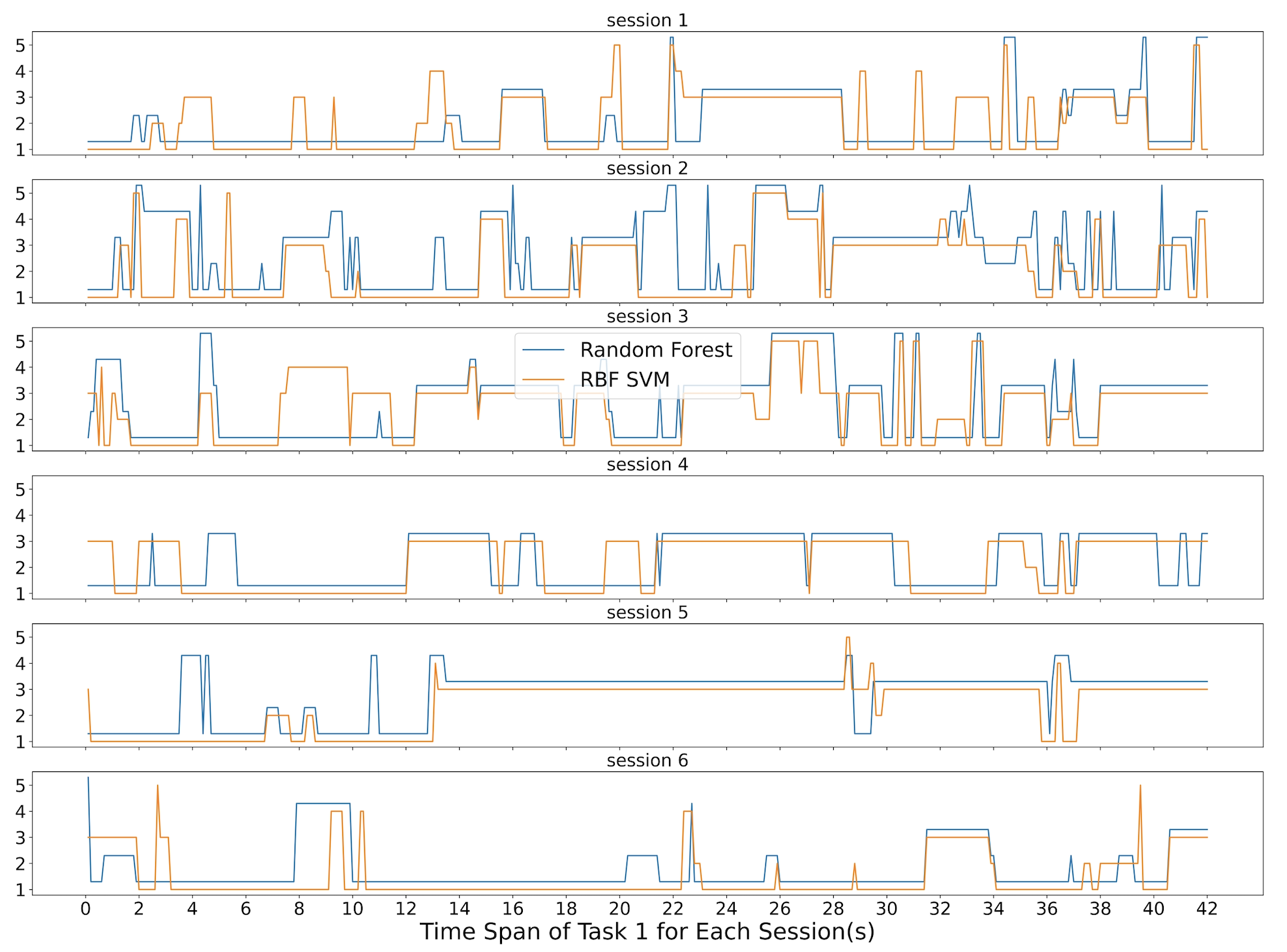}
  \caption{subject 3's task 1's RF and SVM in Phase 1}
  \label{fig:subject_3_task_1_RF_SVM}
\end{figure}
 
\subsection{Identify Noisy Sessions}
 
Figure \ref{fig:subject_3_task_1_RF_SVM} shows what the Random Forest and the RBF SVM in phase 1 predicted during the 42 seconds of subject 3's task 1 for all six sessions. As we can see, both the Random Forest and the RBF SVM mainly produced results with relatively high accuracy in sessions one, four, and six. On the other hand, session two and three had much noise, and session five predicted task 3 for the majority of the time. To minimize the impact of noisy datasets, we calculated the accuracy compared to the ground truth based on the output of Random Forest for each session for each task in phase 1.  

We excluded any sessions that yielded an accuracy of less than 50\%. In the case of Figure \ref{fig:subject_3_task_1_RF_SVM}, we excluded sessions two, three, and five for further machine learning analysis. We reported to subject three and started discussing what may have happened in these sessions. With the exclusion of noisy sessions, the new accuracy for the Random Forest in phase 2 is referred to as "Random Forests Phase 2" later in the paper. We also performed Time Majority Voting on this cleaner dataset. 
 
\subsection{Time Majority Voting}

\begin{table}[!t]
\centering
\begin{tabular} { | c | c | c | c | c| c | c| c | c | c | c |}
\hline
  S/T & t1 RF & \textbf{t1 T} & t2 RF & \textbf{t2 T }& t3 RF & \textbf{t3 T} & t4 RF & \textbf{t4 T} & t5 RF & \textbf{t5 } \\
\hline
    s1 & 0.686 & 0.745 & 0.669 & 0.683 & \textbf{-1} & \textbf{-1} & 0.569 & 0.781 & 0.798 & 0.898\\
\hline
    s2 & \textbf{-1} & \textbf{-1} & \textbf{-1} & \textbf{-1} & \textbf{-1} & \textbf{-1} & 0.693 & 0.838 & 0.633 & 0.707\\
\hline
    s3 & \textbf{-1} & \textbf{-1} & \textbf{-1} & \textbf{-1} & \textbf{-1} & \textbf{-1} & 0.731 & 0.824 & 0.938 & 0.969\\
\hline
    s4 & 0.543 & 0.59 & 0.743 & 0.869 & 0.74 & 0.788 & 0.521 & 0.671 & \textbf{-1} & \textbf{-1}\\ 
\hline
    s5 & \textbf{-1} & \textbf{-1} & 0.645 & 0.738 & 0.588 & 0.8 & 0.593 & 0.743 & 0.662 & 0.829\\ 
\hline
    s6 & 0.762 & 0.871 & \textbf{-1} & \textbf{-1} & 0.507 & 0.714 & 0.812 & 0.917 & 0.681 & 0.845\\ 
\hline
    Average & 0.663 & 0.736 & 0.686 & 0.763 & 0.612 & 0.767 & 0.653 & 0.796 & 0.742 & 0.85\\ 
\hline
\end{tabular}
\vspace{.1in}
\caption{\label{accu_sub_3_RF_TMV} TCR, Accuracy of Top 1 (RF) Phase 2, and TMV(T) (subject 3, by Session(S)/Task(T))}
\end{table}

\begin{figure}[!t]
\centering
  \includegraphics[width=0.91\textwidth]{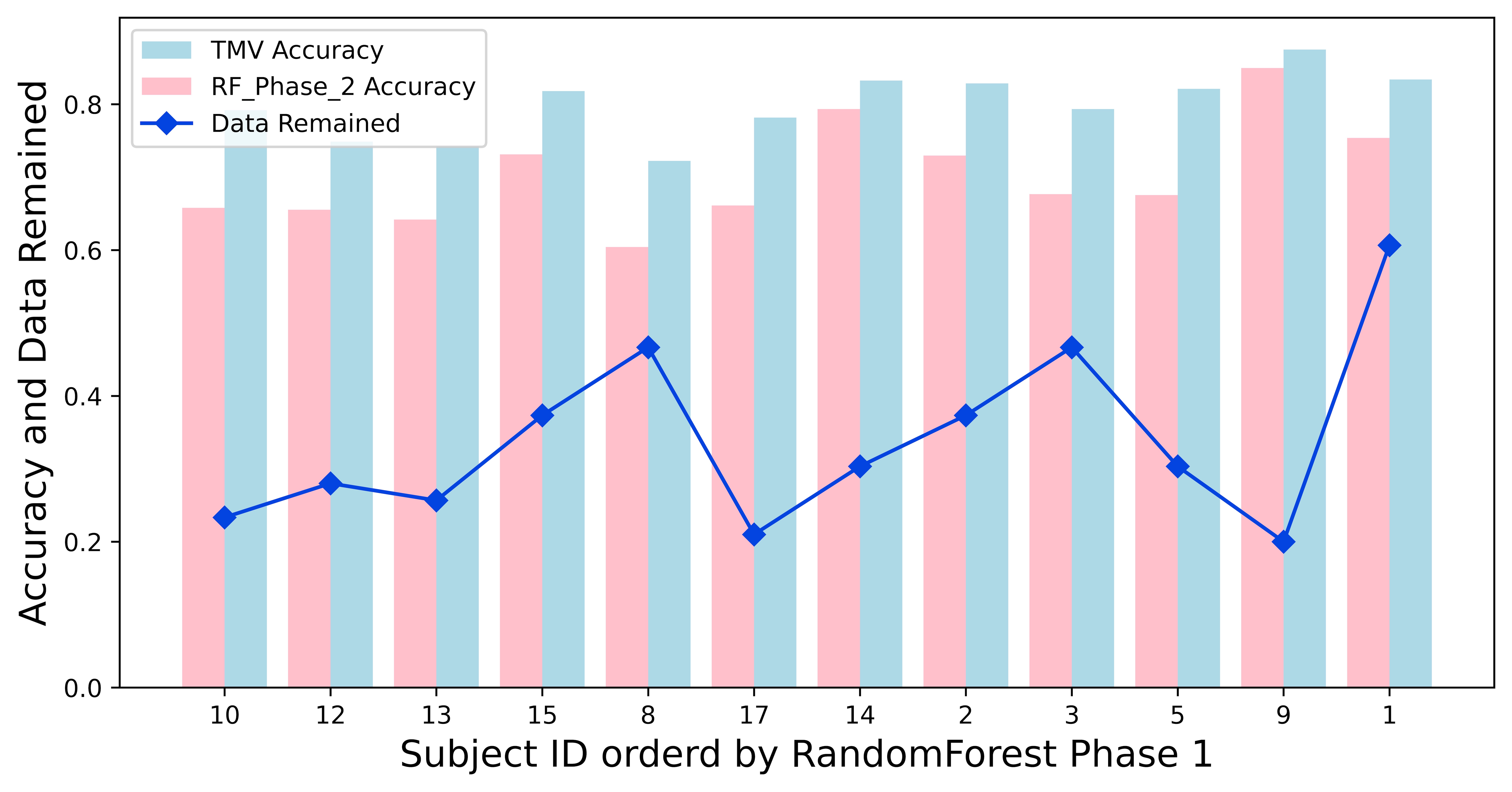}
  \caption{TMV and RF Phase 2}
  \label{fig:TMV_RF_After}
\end{figure}

\begin{figure}[!t]
\centering
  \includegraphics[width=0.99\textwidth]{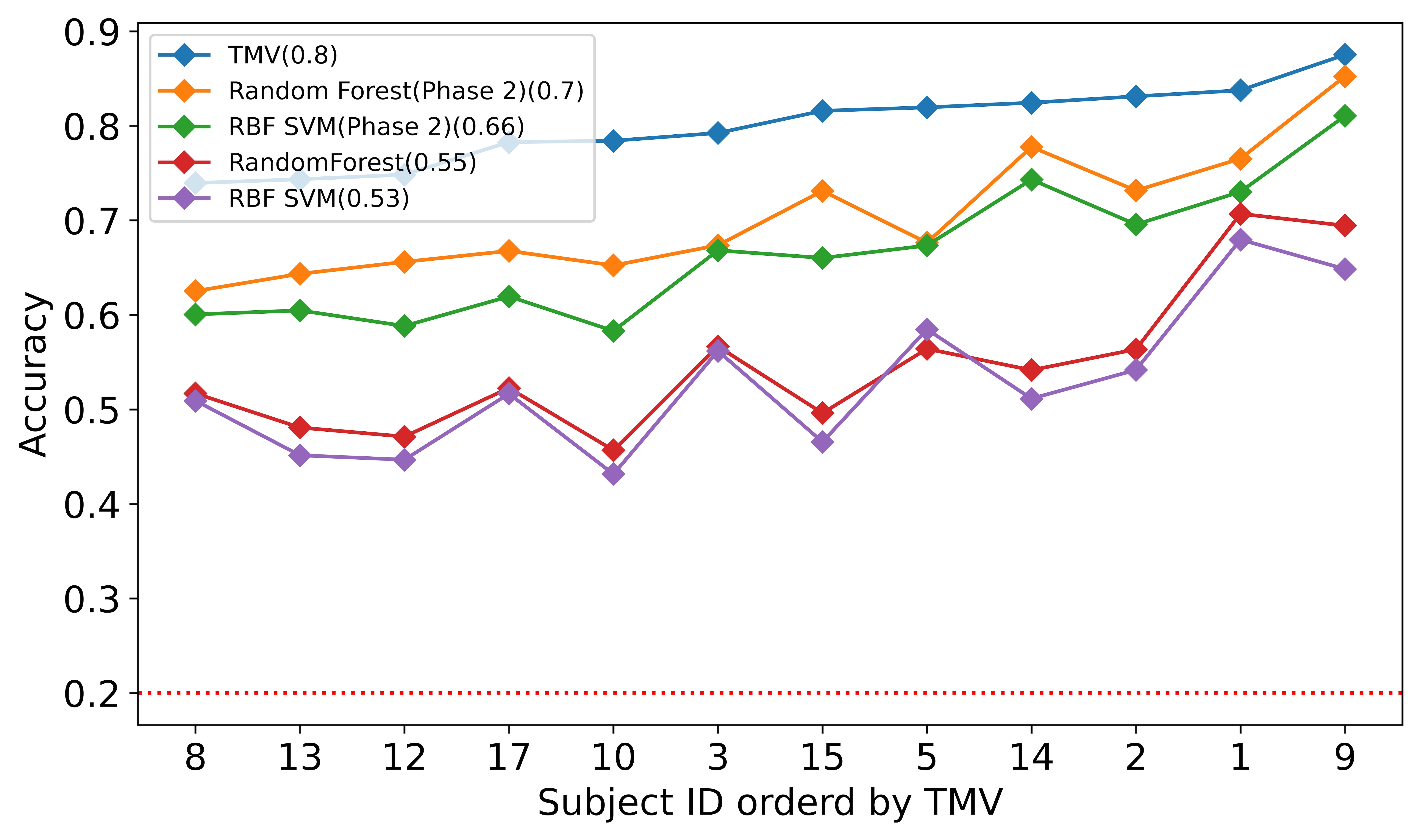}
  \caption{TMV, RF Phase 2, RBF SVM Phase 2, RF phase 1, and RBF SVM Phase 1}
  \label{fig:TMV_algo_compare}
\end{figure}

As shown in Table \ref{accu_sub_3_RF_TMV}, the Time Majority Voting (TMV) has achieved a higher accuracy for subject three, all the six sessions. A value of -1 means that we excluded that session for that task, as we discussed in the previous section. The Random Forest classifier also reached a higher accuracy after cleaning the noisy sessions. As Figure \ref{fig:TMV_RF_After} and Figure \ref{fig:TMV_algo_compare} shows, the pattern is consistent across all the subjects. Figure \ref{fig:TMV_algo_compare} also shows a clear pattern that not only the Random Forest but also the RBF SVM also increased accuracy in phase 2 across all subjects. Table \ref{TMV_RF_runtime} shows the TMV achieved an 80\% average accuracy with an average 74.3 seconds run time. The runtime consists of 39.1 seconds of running the Random Forest and 29.5 seconds of running the RBF SVM. The time for the actual voting is, on average, 5.7 seconds. The training process is the most time-consuming part of this analysis.

\begin{table}[!t]
\centering
\begin{tabular} { | c | c | c |}
\hline
  Algorithms & Average Accuracy & Average code run-time (s) \\
\hline
    \textbf{Time Majority Voting} & \textbf{0.80} & 39.1 + 29.5 + 5.7 = 74.3\\ 
\hline
    Random Forest Phase 2 & 0.7 & 39.1 \\
\hline
    RBF SVM Phase 2  & 0.66 & 29.5 \\ 
\hline
    Random Forest Phase 1 & 0.55 & 39.1 \\
\hline
    RBF SVM Phase 1  & 0.53 & 29.5 \\ 
\hline
\end{tabular}
\vspace{.1in}
\caption{\label{TMV_RF_runtime} TMV and Random Forest with Accuracy and Run-time}
\end{table}

\begin{figure}[!b]
\centering
  \includegraphics[width=0.985\textwidth]{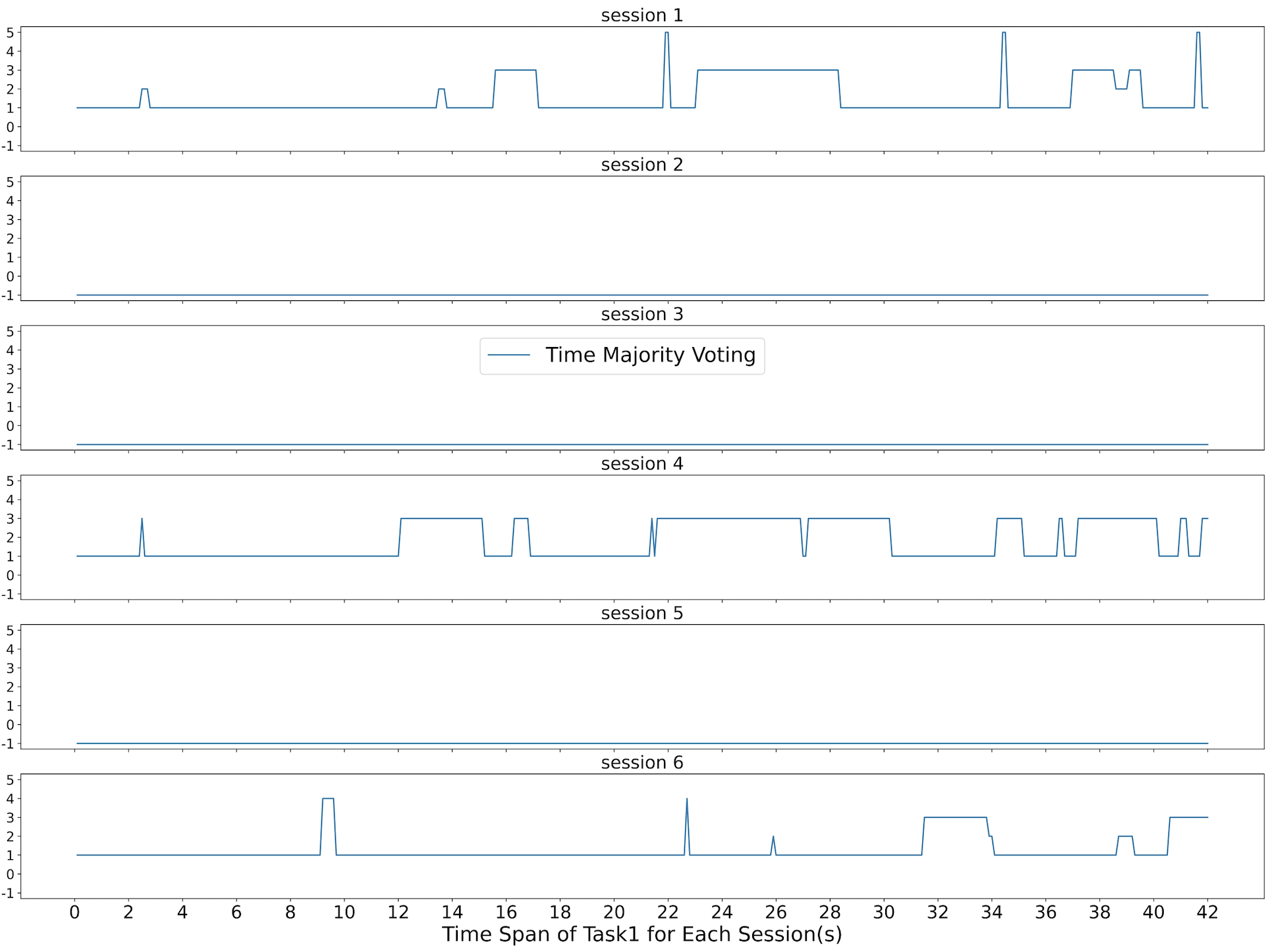}
  \caption{subject 3's task 1's TMV}
  \label{fig:subject_3_task_1_TMV}
\end{figure}

Figure \ref{fig:subject_3_task_1_TMV} shows what the Time Majority Voting(TMV) predicted during the 42 seconds of subject 3's task 1 for all six sessions. As you can see, sessions 2, 3, and 5 have values of -1, which means that they have been excluded. Sessions 1, 4, and 6 have less noise than sessions 2, 3, and 6, and the accuracy is relatively high. 

\section{Discussion}
\subsection{Accuracy and Data Remain} 

The innovation of this method is mainly about temporal dependency. As \cite{lotte2018review, roy2019deep, craik2019deep, qu2018eeg, qu2020identifying} mentioned, EEG data has a significant temporal dependency. The signal of the same task takes about 12 to 18 seconds to switch to the next task. Using majority voting can catch this type of time continuity effect. If both classifiers recognize the same pattern, it is more likely to assure the results. If both of the classifiers recognize the same pattern that is different from the majority result, it is possible that the participants were doing other tasks during the data collection. If only one classifier detects some unusual behaviors, we label it as the majority task of the session. Thus we highlight the noise and keep the remaining data to reflect more on the time continuity nature of the EEG signals.

 \begin{figure}[!b]
\centering
  \includegraphics[width=0.91\textwidth]{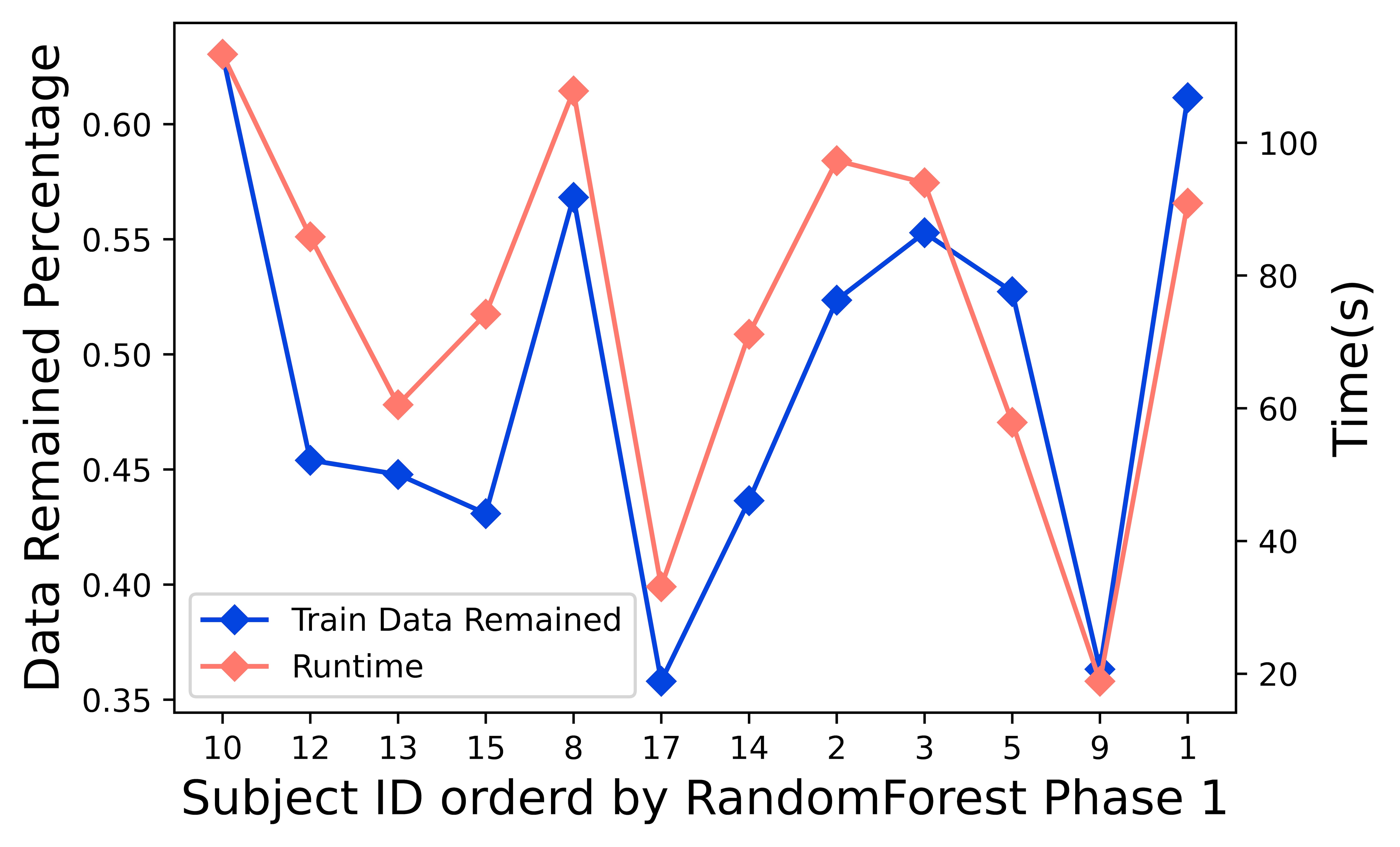}
  \caption{Data Remained for Training and Run-time}
  \label{fig:data_remain_and_run-time}
\end{figure}
 
\subsection{Runtime and Training Data} 
As Figure \ref{fig:data_remain_and_run-time} shows, the run time directly correlated to the training data size. After the data pre-processing, we cleaned up the noise with a plateau longer than a threshold, as mentioned in \cite{lotte2018review, qu2018personalized, qu2020identifying}. We first identified more noisy sessions during the Time Majority Voting process based on the top two classifiers. During this step, more sessions were excluded for further analysis. The training process was the most time-consuming step during the coding running process. Thus the runtime changed together with the size of the training data.

\subsection{Interpretability} 
Figure \ref{fig:subject_3_task_1_RF_SVM} and Figure \ref{fig:subject_3_task_1_TMV} shows two examples of the feedback results we present to the end-users. In this figure \ref{fig:subject_3_task_1_RF_SVM}, the six sessions of subject three, task one is listed as six horizontal charts. The first chart represents session one. Both the Random Forest and the RBF SVM identified the majority task as task one. And for the results of the different predictions, both of the classifiers agreed at some time spots, but not all of them. Our Time Majority Voting algorithm favorite the majority voting results. 

We started with the sessions with good prediction results when demonstrating these figures to each subject. For example, in this figure \ref{fig:subject_3_task_1_RF_SVM}, sessions one, four, and six show pretty consistent patterns. The majority of task prediction results were the designed task one. That implies that the subject may spend more time on task one as planned during these sessions. Session two and three had a lot of different prediction results from both classifiers. Thus we suspected some unexpected reason might cause this situation. We referred back to the experiment notes and reached out to subject three. After discussing with the end-user, we figured out that he had many issues with the sensor signals and was adjusting the EEG headset most of the time during the sessions 2 and 3. Thus we had more information to exclude these sessions from further data analysis. Session five is another situation. The data implied that the subject was doing task three, but the experiment notes were missing for that session, and the subject did not remember the details about that session. Thus, we left a question mark for that session, excluded the session for now, and came up with an improvement plan to keep better experiment notes. This type of machine learning result is explainable to the end-user. 

Such interpretability could contribute to a better understanding of the results and better design for future experiments.

\section{Conclusion}
This paper investigated the state-of-the-art machine learning algorithms that can run on mainstream personal computers for EEG-based BCI. We then proposed a new algorithm, Time Majority Voting (TMV). The results demonstrated that TMV outperformed other existing classifiers. The run time for TMV is still within the acceptable range on a PC. The interpretability of TMV can contribute to a better understanding of the machine learning analysis and an improved design for future experiments.

%
%
%
%
\makeatletter
\renewcommand\@cite[2]{%
  Ref.~#1\ifthenelse{\boolean{@tempswa}}
    {, \nolinebreak[3] #2}{}
}
\renewcommand\@biblabel[1]{#1.}
\makeatother
 \bibliographystyle{splncs04}
 \bibliography{tmv}
\end{document}